\documentclass[conference]{IEEEtran}
\IEEEoverridecommandlockouts
\usepackage{amsmath,graphicx}
\usepackage{amssymb,amsfonts}
\usepackage{cite}
\usepackage{booktabs}
\usepackage{url}
\usepackage{balance}
\usepackage{subfigure}
\usepackage{multirow}
\usepackage[normalem]{ulem}
\usepackage{booktabs}
\usepackage{xcolor}

\graphicspath{{./Figures/}}
\begin{document}
%
\title{StawGAN: Structural-Aware Generative Adversarial Networks for Infrared Image Translation}

%
\author{\IEEEauthorblockN{Luigi Sigillo$^{\ast, \dagger}$, Eleonora Grassucci$^{\ast}$, and Danilo Comminiello$^{\ast}$ 
\thanks{Corresponding author's email: luigi.sigillo@uniroma1.it. This
work has been supported by ``Ricerca e innovazione nel Lazio - incentivi per i dottorati di innovazione per le imprese e per la PA - L.R. 13/2008" under grant number 21027NP000000136, ``Rome Technopole - Progetto Flagship 5" of Sapienza University of Rome under grant number RT52218451F5AAE6, and ``Progetti di Ricerca Grandi" of Sapienza University of Rome under grant number RG11916B88E1942F.}}
\IEEEauthorblockA{$^{\ast}$Dept. of Information Engineering, Electronics and Telecom., Sapienza University of Rome, Italy \\
$^{\dagger}$Leonardo Labs, Rome, Italy}
}


%

\maketitle

\begin{abstract}
This paper addresses the problem of translating night-time thermal infrared images, which are the most adopted image modalities to analyze night-time scenes, to daytime color images (NTIT2DC), which provide better perceptions of objects.
We introduce a novel model that focuses on enhancing the quality of the target generation without merely colorizing it. The proposed structural aware (StawGAN) enables the translation of better-shaped and high-definition objects in the target domain.
We test our model on aerial images of the DroneVeichle dataset containing RGB-IR paired images.
The proposed approach produces a more accurate translation with respect to other state-of-the-art image translation models. The source code is  available at \url{https://github.com/LuigiSigillo/StawGAN}.
\end{abstract}

\IEEEpeerreviewmaketitle

%

%
%
%
%
%
\section{Introduction}
\label{sec:intro}
\begin{figure*}[t]
    \centering
    \includegraphics[width=\textwidth]{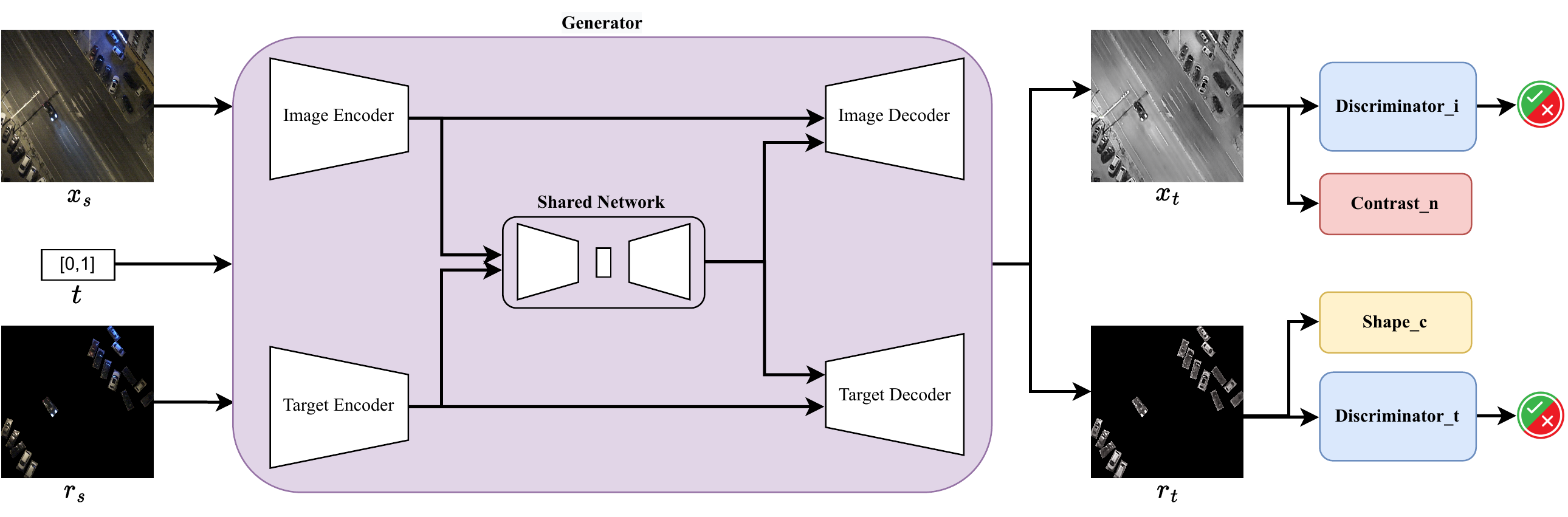}
    \caption{Architecture of StawGAN. We show the part inside of the generator, to explore the shared connections between the two different flows. Indeed this is a translation from RGB to IR, but the same process holds for the contrary.}
    \label{fig:targan_arch}
\end{figure*}
Recently image translation models, methods based on Generative Adversarial Networks (GANs) \cite{NIPS2014_5ca3e9b1}, have shown outstanding performances \cite{9892119}, also in tasks such as colorizing images starting from grayscale or infrared images \cite{10.1007/978-3-030-11024-6_49, 9418801}. GANs are composed of two sub-networks: generator and discriminator. The first generates images in a target domain starting from the input image in the source domain, and the discriminator predicts whether it is a real image or a fake one.
The objective of the training is that the discriminator cannot distinguish anymore between real and generated samples.
Infrared images (IR) are really useful in low-light conditions when standard cameras cannot capture important details like the presence of cars, animals, or pedestrians in a scene of interest. 
IR images, like grayscale images, are treated like single-channel entities, indeed, we focus on thermal infrared (TIR) images, with a very low contrast which is insightful to show hidden details at night, but worsens human perception \cite{Progressincolor}. On the other hand, there are known well-functioning models \cite{9746886, 9880295, Cross-Modality_Paired_Images_Generation, 9496129} that work with RGB images, or at least with the precious annotations provided for those images, for this reason instead of converting those algorithms to the IR field it is interesting to analyze the possibility of translating the images in the RGB domain again. 
Even more recent approaches \cite{wang2022palgan, kumar2021colorization} state that to colorize grayscale images there is no unique solution, this makes this task in the IR domain even more difficult to accomplish since the model has to predict pixels with chrominance but without having luminance information. Existing approaches \cite{zhang2016colorful, zhang2017real} for grayscale images utilize CIE Lab color space instead of RGB, because the number of channels to predict reduce to just two since L, the lightness component, is passed as input and only a and b have to be predicted.
To solve this problem, in \cite{Surez2017InfraredIC} authors decide to split the problem, having three different generators, one for each of the RGB channels to predict. 
Since the translated images are most of the time used for other tasks, like classification or object detection\cite{9418801}, it is of great interest to correctly translate the image details and contours to conduct afterward a better analysis. In a recent work\cite{9703249} authors try to solve the problem using a structured gradient alignment loss ensuring edge consistency for the translated images, though to do this they exploit the usage of canny edges\cite{Yang2014MultifeatureBasedSI}.

The aim of this paper, instead, is to intervene with a novel full-neural approach in the generation of meaningful images, with only one generator, for both IR and RGB images, enhancing the quality of possible targets (i.e. bus, car, etc.) for a task of interest to be accomplished. Inspired by the TarGAN\cite{Chen2021TarGANTG}, an image modality translation model used for medical image synthesis, we propose the structured-aware generative adversarial networks (StawGAN). One of the most important properties of our model is that we enforce a better generation of the targets using different flows in the same model for the image and the segmented target, with the flows being interconnected by a shared architecture. We make the network target- and structural-aware by optimizing a Structural Dissimilarity Index Measure (DSSIM) loss encouraging spatial smoothness in the generated images. 
Our method lies in the middle between paired and unpaired image translation models, because it is generalizable to unpaired problems, but can also be used in paired ones taking advantage of \textit{ad hoc} losses. To prove the effectiveness of our method, we conduct several experiments for image modality translation on the aerial images of the DroneVehicle\cite{9759286} dataset. Our method improves the state-of-the-art in most of the metrics we use to compare to other models, thus proving that the proposed approach enhances the generation of well-shaped targets.

The paper is organized as follows. Section~\ref{sec:novelty} introduces the model proposed, Section~\ref{sec:exp} reports the experimental evaluation, while we draw conclusions in Sec.~\ref{sec:conclusion}.



%
\section{Proposed Method}
\label{sec:novelty}
To introduce our approach, we briefly review the recent Target-Aware Generative Adversarial Network (TarGAN) \cite{Chen2021TarGANTG}. This network translates images from a source domain to a target domain while paying particular attention to the region of interest, passed as a segmentation map during the translation, this model is originally made for a medical scenario where this attention to the details can be crucial. The input data for the generator model are the image in the source domain ($s$), the segmented image, and the one-hot vector representing the target domain ($t$), the output will be the translated image and its segmentation:
\begin{equation*}
G(x_s, r_s, t) \rightarrow (x_t, r_t)
\end{equation*}

\noindent The key to enhancing the generation of vehicles lies in the shared connections between the two flows of the model, composed of two encoder-decoder architectures, one for the whole image and another for the segmented one, while between the two flows there are shared layers to combine information for a better translation. 
As the original TarGAN, we involve a Shape Controller $S$ that takes as input the fake target image $r_t$ and generates a binary mask representing the foreground area of the generated images.
Additionally, we introduce a novel Contrast Network $C$, to help the generator in choosing the right contrast, sharpness, and gamma factors.

For training, we employ adversarial loss, which is the canonical loss to train adversarial networks, and a modality classification loss to force the target modality. It is applied on the generator $G$, on the discriminator $D_x$, and on the discriminator $D_r$. 
$\mathcal{L}_{{\text{cls}}_{-} x}^{r}$ refers to the loss of real images, thus used for the discriminators, while $\mathcal{L}_{{\text{cls}}_{-} x}^{f}$ to the fake ones and used on the generator $G$. Since some losses are the same for the image and the target, in the next equations, $z$ could represent an image $x$ or a target image $r$:
\begin{equation}
    \begin{aligned}
    \mathcal{L}_{{\text{cls}}_{-} z}^{r} &=\mathbb{E}_{z_{s}, s}\left[-\log D_{{\text{cls}}_{-} z}\left(s \mid z_{s}\right)\right]\\
    &+ \lambda_{u} \mathbb{E}_{z_{t}, s^{\prime}}\left[-\log D_{{\text{cls}}_{-} z}\left(s^{\prime} \mid z_{t}\right)\right],
    \end{aligned}
\end{equation}


\begin{equation}
    \mathcal{L}_{{\text{cls}}_{-} z}^{f}=\mathbb{E}_{z_{t}, t}\left[-\log D_{{\text{cls}}_{-} z}\left(t \mid z_{t}\right)\right].
\end{equation}

\noindent We use the shape consistency loss\cite{Zhu2019UGANUG} since the structural shape of original images could sometimes slightly differs from one of the translated samples, so the shape loss tries to contain this behavior by:
\begin{equation}
\mathcal{L}_{\text {{\text{shape}}}_{-} r}=\mathbb{E}_{r_{t}, b^{x}}\left[\left\|b^{x}-S\left(r_{t}\right)\right\|_{2}^{2}\right].
\end{equation}

We introduce the structural information in the model by involving the Structural Dissimilarity Index Measure (DSSIM) \cite{1284395} in the optimization process of the proposed network, so to improve the estimation of both the luminance and chrominance pixels of TIR images. The DSSIM loss defines a region, called window size, where we are predicting the luminance and chrominance of the image, and it is defined as: 
\begin{equation}
    \mathcal{L}_{{\text{ssim}}} = \frac{1-\text{SSIM}(x_{{\text{paired}}}-x_t)}{2},
\end{equation}
where $x_{{\text{paired}}}$ is the original image but in the same domain of $x_t$, which is instead the translated sample.
Since we apply also the reconstruction loss \cite{8237506} aims at preserving the original characteristics of the source images, we employ it with the DSSIM loss as suggested in \cite{9703249}. We are using the reconstruction for both the target and the entire image, while DSSIM is only for the latter:

\begin{equation}
\label{eq:l_rec}
\mathcal{L}_{{\text{rec}}_{-} z}=\mathbb{E}_{z_{s}, z_{s}^{\prime}}\left[\left\|z_{s}-z_{s}^{\prime}\right\|_{1}\right] +\lambda_{{\text{ssim}}} \mathcal{L}_{{\text{ssim}}}
\end{equation}

\noindent We employ crossing loss to force the generator to pay attention to the target area when generating the translated image. Denoting with $y$ the target area label of $x_s$, we have:
\begin{equation}
    \mathcal{L}_{\text {cross }}=\mathbb{E}_{x_{t}, r_{t}, y}\left[\left\|x_{t} \cdot y-r_{t}\right\|_{1}\right],
\end{equation}

To train the Contrast Network and the Generator, we involve the $L_1$ loss that takes the fake image, with the applied factors, and the paired image, serving in reducing the production of more gray-like and more colorful images\cite{Isola2017ImagetoImageTW}.
The complete objective function takes therefore the following form, with hyperparameters lambda to control the weights of the loss functions:

\begin{equation}
\begin{aligned}
\label{eq:complete_obj_tar}
    \mathcal{L}_{D_{(z)}} &=-\mathcal{L}_{{\text{adv}}_{-}(z)}+\lambda_{{\text{cls}}}^{r} \mathcal{L}_{{\text{cls}}_{-}(z)}^{r}, \\
    \mathcal{L}_{G} &=\mathcal{L}_{{\text{adv}}_{-}(z)}+\lambda_{{\text{cls}}}^{f} \mathcal{L}_{{\text{cls}}_{-}(z)}^{f}+\lambda_{\text{rec}} \mathcal{L}_{{\text{rec}}_{-}(z)}  \\
            &+\lambda_{{\text{cross}}} \mathcal{L}_{{\text{cross}}} 
            +\lambda_{{L}_1}\mathcal{L}_{{L}_1}, \\
 \mathcal{L}_{G, S} &=\mathcal{L}_{{\text{shape}}_{-}(r)},\\
 \mathcal{L}_{C} &=\lambda_{{L}_1}\mathcal{L}_{{L}_1}.\\
\end{aligned}
\end{equation}

\noindent Therefore, thanks to the DSSIM loss in \eqref{eq:l_rec}, the proposed model is aware of structural information producing better-shaped objects.

The encoder-decoder couples in the generator are composed of different building blocks: the convolutional one, with the usage of convolution, instance normalization, and Leaky ReLU, and the upsampling one involving the same operation as the previous one but with an upsampling operation done by interpolation at the beginning. 
For the discriminator, differently from the previous model, we are using two identical models but as separate entities, one determines if an image is fake or real and the other does the same, but for the segmented one, furthermore being a conditional GAN, they determine even if the domain is being correctly translated. After each convolutional layer of the discriminators, we adopted a layer of spectral normalization \cite{miyato2018spectral} to further stabilize training. The only difference between the two discriminators regards the final layer. The first one involves kernel size $3$, and stride and padding $1$. The other one, instead, varies the dimension depending on the number of target modalities and the kernel depending on the image size.
The Shape Controller, instead, is built with convolutional blocks introduced before, while different from the discriminator the upsampling is done via transposed convolutions. Finally, the Contrast Network is composed of a convolutional layer, a max pooling, another convolutional layer, and finally three fully connected layers. We pass the fake image in this model, that predicts the desired values mentioned above. We show an overview of the complete architecture in Fig.~\ref{fig:targan_arch}

\section{Experimental evaluation}
\label{sec:exp}
\subsection{Dataset}
To test our image modality translation task we utilize the DroneVehicle\cite{9759286} dataset. It contains $56,878$ samples taken by drones divided between half RGB and half infrared image. The data comes with annotations with oriented bounding boxes, labeled with five different categories ordered respectively by a number of occurrences in the dataset: car, truck bus, van, and freight car. The image size is $840 \times 712$. The difference between RGB and IR images is in the number of channels for our analysis, considering the IR as grayscale images with only one channel. For instance, IR images are divided into near-infrared, closer to the visible spectrum, and thermal infrared, which is useful in dark light conditions of the emitted energy from objects. The dataset we consider is composed of the latter type of IR. For our experiments we employ all the data resizing images at $256 \times 256$, obtaining approximately $17000$ samples for the training dataset, and $3000$ for the validation one. The only preprocess we do is the removal of the darkest samples, around $2000$ samples. The images are normalized between -1 and 1. For our analysis is essential to have the segmentation masks of the targets, for this aim we built a dataset of masks starting from the bounding boxes provided by the original dataset. The segmentation masks are divided also into different colors depending on the class of the target, it is not of interest to our model, but can be for future work involving this dataset. 
\subsection{Metrics}
To perform an objective evaluation, we compute several metrics: the Fréchet inception distance (FID)\cite{heusel2017gans}, the Inception Score (IS)\cite{NIPS2016_8a3363ab}, the structural similarity index (SSIM)\cite{1284395}, and the PSNR. FID estimates how the distributions of real and reconstructed images are far from each other, and IS assesses how realistic the generated images are by measuring the images' variety and truthfulness. SSIM measures image degradation through structural information.
To evaluate the results of segmentation instead, we use the Dice similarity coefficient (DSC)\cite{10.2307/1932409}, the Segmentation score (S-Score) \cite{Zhang2018TranslatingAS}, and the mean absolute error (MAE). DSC quantifies how much two sets are similar averaging the size of their intersection. 
S-score sees how the shape of a segmentation behaves after the translation in another domain.
For each fake segmented image, S-score is computed with the DSC between the generated segmented image and the ground truth of the real image. $DSC(SegNet(s), g)$ where $s$ is the generated image, and $g$ is the original image. Here $SegNet$ is the segmentation part of the generator, so given in input the yet translated image we are taking only the segmented image as output and then evaluating with the ground truth segmented image.
\begin{figure*}[t]
    \centering
    \includegraphics[width=\textwidth]{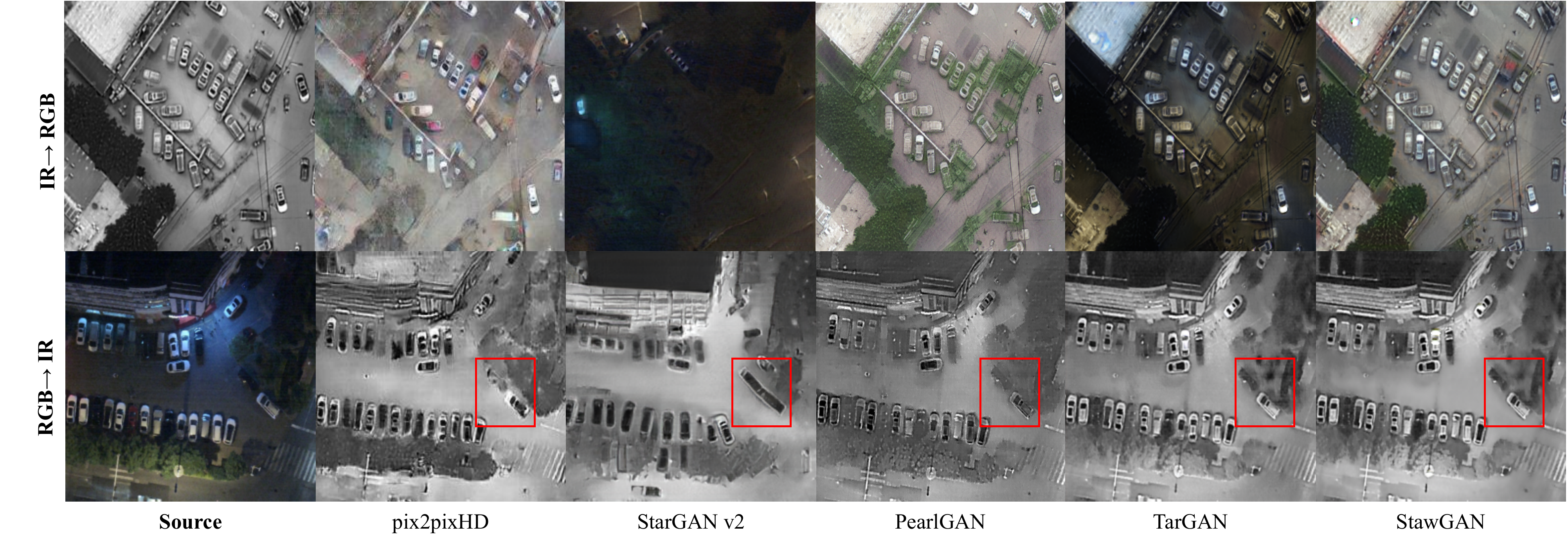}
    \caption{Random samples of StawGAN image modality translations on the DroneVehicle dataset. The first column refers to the original domain images, IR in the first row, and RGB in the second. The corresponding translations in the RGB domain are in the first row and subsequent columns, while IR is in the second one. Our approach produces better samples in both domains with respect to other state-of-the-art models.}
    \label{fig:targan_results}
\end{figure*}
\subsection{Experimental Setup}
StawGAN is implemented with PyTorch and trained on an NVIDIA Tesla V100 GPU, experiments last two and a half days. We tune different hyperparameters to successfully train this model. We set the lambda of the cycle loss $\lambda_{{\text{cycle}}}=10$, the one of the cross loss to $\lambda_{{\text{cross}}}=25$,
while all the others are equal to 1. We start to involve the DSSIM loss only after epoch 10. There are different learning rates for the discriminators and the other networks. The optimizer we use is Adam\cite{DBLP:journals/corr/KingmaB14} with $(\beta_1, \beta_2) = (0.5, 0.9)$. The learning rate for the discriminator is $0.0003$ while for the other models is $0.0002$, both during the training start to decrease linearly to zero after half of the epochs, which is set to 50. The batch size is set to 12. Due to the faster convergence of the target flow of the network, we train it only during even epochs number.

\subsection{Results and Discussion}
We compare our method with different models that reached state-of-the-art performance on other datasets. We test pix2pixHD\cite{Wang2018HighResolutionIS} used for unpaired image-to-image translation, StarGAN v2\cite{Choi2020StarGANVD} interesting for the characteristics of extracting the style code from the scene and not injecting it with a fixed label, PearlGAN\cite{9703249} introduced for this specific task of translation from IR to RGB and finally the original version of TarGAN. 
Figure~\ref{fig:targan_results} shows the comparison on randomly chosen samples.
For the translation $IR\rightarrow RGB$ the results of PearlGAN are visually poor considering because of the green color of the buildings, supposing that it interprets them as vegetation. 
Regarding the translation $RGB\rightarrow IR$ the results of StarGAN v2 are slightly blurred, mostly on the targets, PearlGAN does not translate correctly all the cars, and some of them are ghosted. TarGAN and pix2pixHD samples are visually comparable to our results even though our method produces more realistic colors, but still, if we focus on the red boxes in Fig.\ref{fig:targan_results}, we notice a concrete difference. From RGB to IR pix2pixHD loses the details of the target car, while TarGAN shows black holes and paler. For the opposite translation, both the color and the shape of the StawGAN image are more appealing with respect to pix2pixHD. Furthermore, TarGAN loses information on the targets, since the image is darker, making it difficult to distinguish the shape and the type of vehicle. In the end, the proposed StawGAN produces high-definition samples with realistic colorization and well-shaped and defined targets. Table~\ref{tab:res_reference} reports objective metrics results for the image modality translation task. From this analysis, we can confirm that the FID of pix2pixHD is higher with respect to our approach that scores the second-best value, but our method surpasses it in all the other metrics.
Those results confirm our theoretical statements, and prove the effectiveness of the proposed method with respect to the state of the art, clearly showing the significance of the proposed StawGAN.


\begin{table}[ht]
\scriptsize
\centering
\caption{Quantitative comparison on image modality translation.}
\label{tab:res_reference}
\begin{tabular}{|l|c|c|c|c|}
\toprule
Model & FID$\downarrow$ & IS$\uparrow$ & PSNR$\uparrow$ & SSIM$\uparrow$ \\
\midrule
pix2pixHD\cite{Wang2018HighResolutionIS} & $\underline{0.0259}$	&	$\mathbf{4.2223}$	&	$11.2101$	&	$0.2125$	  \\
StarGAN v2\cite{Choi2020StarGANVD} & $0.4476	$	&$2.7190$		&$11.2211$&		\underline{$0.2297$}	 \\
PearlGAN\cite{9703249}& $0.0743$ & $\underline{3.9441}$ &		$10.8925$	&	$0.2046$	\\
TarGAN\cite{Chen2021TarGANTG}& $0.1177$	&	$3.4285$	&	\underline{$11.7085$}	&	$0.2382$	\\
StawGAN & $\mathbf{0.0119}$	 &	$3.5163$	&	$\mathbf{11.8251}$	&	$\mathbf{0.2453}$	 \\
\bottomrule
\end{tabular}
\end{table}

\subsection{Segmentation}
We perform also a secondary analysis on the segmentation capability of the proposed model with respect to the TarGAN, which is the only model from the one we compare that performs segmentation too. The comparison shows that our model gains better performance with respect to the TarGAN. Since segmented images have in both domains large black zones, the translation task results easier to be accomplished. Therefore, the segmentation metrics are evaluating mainly the generation of concrete target shapes. In Tab.~\ref{tab:res_segmentation} we report objective metrics results for this task.

\begin{table}[t]
\scriptsize
\centering
\caption{Quantitative comparison on segmentation.}
\label{tab:res_segmentation}
\begin{tabular}{|l|c|c|c|c|}
\toprule
Model & DSC$\uparrow$ & S-Score$\uparrow$ & MAE$\downarrow$ \\ 
\midrule
TarGAN\cite{Chen2021TarGANTG}& $79.34$	&	$85.08$&	$0.0115$	\\ 
StawGAN & $\mathbf{84.27}$ &	$\mathbf{88.61}$ &	$\mathbf{0.0081}$	\\ 
\bottomrule
\end{tabular}
\end{table}
%
%
%
%
%
\section{Conclusion}
\label{sec:conclusion}
In this paper, we introduced the StawGAN, an image translation model that enhances the synthesis of images in another domain by being structurally consistent. The proposed method generates from an input image the corresponding image in IR or RGB domain using also masked targets, exploiting the shared connections in the generator, improving the resultant translation. This neural approach strengthens edges in the translation, making it useful for a variety of sub-tasks like object detection or classification. Moreover, we can use this method with both paired or unpaired datasets.

%
%
\balance
\bibliographystyle{IEEEbib}
\bibliography{QWavelet}
\end{document}